\documentclass[journal]{IEEEtran} 

\IEEEoverridecommandlockouts                              

\usepackage[utf8]{inputenc}
\usepackage[T1]{fontenc}

\usepackage{graphicx} 
\usepackage{epsfig} 
\usepackage{mathptmx} 
\usepackage{mathptmx} 
\usepackage{amsmath} 
\usepackage{amssymb}  
\usepackage{subfigure}
\usepackage[dvipsnames]{xcolor}
\usepackage{float}
\usepackage{todonotes}
\usepackage{multirow}
\usepackage{cite}
\usepackage{xcolor}
\usepackage{bm}
\usepackage{hhline}
\usepackage[mathscr]{euscript}
\usepackage{fancyhdr}
\usepackage{tikz}
\usepackage{lipsum}

\newcommand{\B}[1]{\bm{#1}}

\newcommand\copyrighttext{%
\footnotesize This work has been submitted to the IEEE for possible publication. Copyright may be transferred without notice, after which this version may no longer be accessible.}
\newcommand\copyrightnotice{%
\begin{tikzpicture}[remember picture,overlay]
\node[anchor=south,yshift=10pt] at (current page.south) {\fbox{\parbox{\dimexpr\textwidth-\fboxsep-\fboxrule\relax}{\copyrighttext}}};
\end{tikzpicture}%
}

\usepackage{tikz}
\title{\LARGE \bf Efficient Force Estimation for Continuum Robot}

\author{Qingyu~ Xiao$^{1}$,
        and Yue~Chen$^{1}$
\thanks{This research was partially funded by Georgia Institute of Technology Startup Grant. Corresponding author: Yue Chen}
\thanks{$^{1}$Qingyu Xiao and Yue Chen are  with the Department of Biomedical Engineering, Georgia Institute of Technology/Emory University, Atlanta, GA, 30332.
        {\tt\small qxiao33@gatech.edu; yue.chen@bme.gatech.edu}}%
}

\usepackage{xcolor}
\begin{document}
\maketitle
\begin{abstract}
External contact force is one of the most significant information for the robots to model, control, and safely interact with external objects. For continuum robots, it is possible to estimate the contact force based on the measurements of robot configurations, which addresses the difficulty of implementing the force sensor feedback on the robot body with strict dimension constraints. In this paper, we use local curvatures measured from fiber Bragg grating sensors (FBGS) to estimate the magnitude and location of single or multiple external contact forces. A simplified mechanics model is derived from Cosserat rod theory to compute continuum robot curvatures. Least-square optimization is utilized to estimate the forces by minimizing errors between computed curvatures and measured curvatures. The results show that the proposed method is able to accurately estimate the contact force magnitude (error: 5.25\% -- 12.87\%) and locations (error: 1.02\% -- 2.19\%). The calculation speed of the proposed method is validated in MATLAB. The results indicate that our approach is 29.0 -- 101.6   times faster than the conventional methods. These results indicate that the proposed method is accurate and efficient for contact force estimations. 


\end{abstract}

\copyrightnotice

\section{Introduction}
\IEEEPARstart{C}{ontinuum} robots could easily change their configurations due to external contact, which will affect the robot tip motion and eventually alter the control efforts. Many work has been done to estimate the external contact forces. Contact force can be directly measured by embedding miniature force sensors on the soft robot body \cite{mitsuishi2007force,valdastri2006integration}. However, these methods are only effective when the contact occurs at the sensor location, for instance at the tip of the robot. Also, the strict dimension constraints also limit the wide application of these approaches, especially in the minimally invasive surgical tools where the tool tip often equipped with multiple treatment or diagnosis units \cite{alipour2019mri}. Alternatively, continuum robot contact force can be estimated based on actuator space measurement.  For example, Bajo et al \cite{{bajo2010finding,bajo2011kinematics}} proposed a method for tendon-driven continuum robot contact force sensing, but this requires dedicated force cell at the actuation unit. Also, the actuation-space force cells may not be available for other continuum devices such as concentric tubes, steerable needles,  catheters, or guidewires. 


Alternatively, force can be estimated implicitly based on the measurements of continuum robot configuration change. This is applicable since the intrinsic compliant nature of continuum robots allows the robot configuration to be changed with the contact forces \cite{webster2010design}. Numerous models have been published to predict the continuum robot configurations subjected to external loads, including beam-based method \cite{stilli2018static}, calibration method \cite{wang2019geometric}, variable curvature method \cite{mahl2014variable}, and Cosserat rod theory \cite{rucker2010geometrically}. 
Contact forces, in turn, can be estimated by matching the measured configuration with modelled data. 


Many shape-based methods were proposed to estimate the contact forces. Prior research used shape-based method to estimate the tip force  \cite{{rucker2011deflection,khoshnam2015modeling,khan2017force,hasanzadeh2015model,back2016new}}, but they were unable to estimate the forces applied along the body of continuum robot. 
Recent publications have shown the feasibility of using shape-based method to estimate the contact force along the body of the continnum robot. Qiao \cite{qiao2021force} proposed a shape-based method which can detect the point forces (mean error of 15.4\%) and locations (mean error of 6.51\%). Aloi \cite{aloi2019estimating} proposed a shape-based method based on Fourier transformation. This method will estimate the force in a distributed fashion and can detect the force location by observing the "peaks" of the distributed force profile. However, it cannot provide useful information to the exact force magnitude and location if it is point contact.

Aside from measuring the shape of continuum robot, curvature, which is invariant under rotation, can provide more robust information for force estimation. This is because the curvature will not change for any rotations the continuum robot undergoes. Thus, the curvature-based methods are still valid even if the robot orientation is not well-calibrated.  The curvatures along the continuum robot can be measured using fiber Bragg grating sensors (FBGS) \cite{gouveia2011temperature,xu2016curvature,ge2016bidirectional,li2019distributed,barrera2015multipoint}.  Several curvature-based methods have been investigated for force estimation. Qiao \cite{qiao2019estimating} utilized FBGS to measure the curvatures along the manipulator and calculates the force (mean error of 9.8\% error) and locations (mean error of 3.6\% error) directly using constitutive law. Al-Ahmad \cite{al2021fbg} utilized unscented Kalman filter to the measured curvatures, which achieved 7.3\% error of force magnitude and 4.6\% error of location.

In this paper, we present a new curvature-based force estimation method by representing the Cosserat rod theory in local frame to enable faster curvatures computation, which avoid the time-consuming integration of rotation matrix. Based on the simplified model, least squares optimization is adopted to estimate the contact forces in a significantly efficient approach. The contributions of this paper include:
\begin{enumerate}
    \item Analyzed the conditions for accurate force estimation via curvatures.
    \item Derived a simplified model for fast curvature computation and force estimation.
    \item Experimentally validated the proposed force estimation method on single, double and triple force cases.
\end{enumerate}
The structure of the paper is arranged as follows. Section II describes the derivation of a simplified model from Cosserat rod theory to compute curvatures. Section III discusses the conditions for pure curvature-based force estimation method and the estimation procedures. Section IV shows the experimental results of the force estimation regarding accuracy and computation time. Finally, Section V is the conclusion of this paper.

\section{Curvature Calculation}
\subsection{Review of Cosserat Rod Model}
Cosserat rod model describes the equilibrium state of a small segment of thin rod subjected to internal and external distributed forces as well as internal and external distributed moments. Previous works \cite{jones2009three} have bridged the differential geometry of a thin rod with Cosserat rod theory so that the deformation of the rod can be estimated by solving ODEs in (\ref{eq:ode1}) - (\ref{eq:ode4}) given external forces $\B F_{e}$ and external moments  $\B L_e$. To distinguish variables from different frames, we use lowercase letters for variables in local frame and uppercase letters for variables in global frame.
\begin{eqnarray}
\dot{\B P} &=& \B R \B v \label{eq:ode1}\\
\dot{ R} &=& R \hat{\B u} \label{eq:ode2}\\
\dot{\B N} &=& -\B F_e \label{eq:ode3}\\
\dot{\B M} &=& -\dot{\B P}\times \B N - \B L_e \label{eq:ode4}
\label{eq:Cosserat ODEs}
\end{eqnarray}
where the dot symbol in $\dot{\B x}$ represents the derivative of $\B x$ respect to the arc length $s$, hat symbol in $\hat{\B x}$ reconstruct vector $\B x$ to a 3 by 3 skew symmetric matrix, $\B P$ is the shape in global frame, $R$ is the rotation matrix of local frame relative to the global frame, $\B N$ and $\B M$ are internal force and moment in global frame which obey the constitutive law 
\begin{eqnarray}
\B N &=& RK_{SE}(\B v-\B v^*)\label{eq:consitituitive law1}\\
\B M &=& RK_{BT}(\B u-\B u^*)
\label{eq:consitituitive law2}
\end{eqnarray}
$\B v$ and $\B u$ are differential geometry parameters of a rod, $\B v^*$ and $\B u^*$ are differential geometry parameters of a rod with no external loads. $K_{SE}$ and $K_{BT}$ are the stiffness matrices. The boundary conditions for solving the ODEs (\ref{eq:ode1}) - (\ref{eq:ode4}) are described as
\begin{eqnarray}
\B N(loc_{tip}) = \B F_{tip}\\
\B M(loc_{tip}) = \B T_{tip}
\end{eqnarray}
where $\B F_{tip}$ and $\B T_{tip}$ are external force and torque applied at the tip. $loc_{tip}$ is the tip position in arc length. Thus, the rod shape $\B P$ can be computed by solving the boundary value problem (BVP).
\subsection{ODEs For Curvature Calculation}
The widely accepted Cosserat rod model in (\ref{eq:ode1}) -- (\ref{eq:ode4}) are expressed in global frame. But it is beneficial to use the model in local frame to calculate the curvatures. We firstly take the derivative of $\B M$ in (\ref{eq:consitituitive law2}) and combine with (\ref{eq:ode4})
\begin{equation}
    R \hat{\B u}K_{BT}(\B u-\B u^*) + RK_{BT}\dot{\B u} = -\dot{\B P}\times \B N-\B L_e
\end{equation}
Then, using the relation $\dot{\B P}\times \B N = \hat{\dot{\B P}}\B N$, $\dot{\B P} = R\B v$, and $\hat{(R\B v)} = R\hat{\B v}R^T$ (this is true when $R \in SO(3)$), we can obtain
\begin{equation}
      R \hat{\B u}K_{BT}(\B u-\B u^*) + RK_{BT}\dot{\B u} = -R\hat{\B v}R^T \B N-\B L_e
\end{equation}
Multiplying $R^T$ on both side, and solve for $\dot{\B u}$, we can have
\begin{equation}
    \dot{\B u} = -K_{BT}^{-1}\left( \hat{\B u}K_{BT}(\B u-\B u^*) + \hat{\B v}R^T\B N + R^T\B L_e \right) \label{eq:udot_local1}
\end{equation}

We define local variables $\B n = R^T\B N$, and $l_e = R^T\B L_e$. The meaning of the variables $\B n$ and $\B l_e$ are actually the variables $\B N$ and $\B L_e$ expressed in local frame, respectively. Hence, (\ref{eq:udot_local1}) becomes
\begin{equation}
    \dot{\B u} = -K_{BT}^{-1}\left( \hat{\B u}K_{BT}(\B u-\B u^*) + \hat{\B v}\B n + \B le \right)
\end{equation}
In order to solve $\B n$, we use (\ref{eq:ode3}) and multiply $R^T$ on the two sides and add $\dot{R}^T\B N$, which gives
\begin{equation}
    R^T\dot{\B N} + \dot{R}^T\B N = -R^T\B Fe + \dot{R}^T\B N\label{eq:ndot_local1}
\end{equation}
Again, we define $R^T\B F_e = \B f_e$. Notice that the left hand side in (\ref{eq:ndot_local1}) is actually the derivative of $R^T\B N$, which is  $\dot{\B n}$. $\dot{R}$ can be replaced by (\ref{eq:ode2}) on the right hand side. Thus we have
\begin{equation}
    \dot{\B n} = -\B fe+\hat{\B u}\B n 
\end{equation}

To sum up, the curvatures of a general rod can be calculated by 
\begin{eqnarray}
    \dot{\B u} &=& -K_{BT}^{-1}\left( \hat{\B u}K_{BT}(\B u-\B u^*) + \hat{\B v}\B n + \B le \right) \label{eq:ode_local1}\\
    \dot{\B n} &=& -\B fe + \hat{\B u}\B n \label{eq:ode_local2} 
\end{eqnarray}
with boundary conditions
\begin{eqnarray}
\B u(loc_{tip}) &=& \B 0 \label{eq:boundary_u=0}\\
\B n(loc_{tip}) &=& \B f_{tip}
\end{eqnarray}
where $loc_{tip}$ and $f_{tip}$ are the location and applied external force at the tip, respectively. (\ref{eq:ode_local1}) and (\ref{eq:ode_local2}) can be simplified in a special case where the rod is straight ($\B u^* = \B 0$ and $\B v^* = [0\quad0\quad 1]^T$), with circular cross-section ($K_{11} = K_{22} = K_{33}/2 $), nonshear and inextensible ($\B v  = [0\quad0\quad 1]^T$), and no external moment ($\B l_e=\B 0$). The simplified ODEs can be written as

\begin{eqnarray}
\dot{u}_x &=& -n_y/K_{BT,11} \label{eq:ode_local_simp1}\\
\dot{u}_y &=& n_x/K_{BT,22}\label{eq:ode_local_simp2}\\
\dot{n}_x &=& -f_x + u_y n_z \label{eq:ode_local_simp3}\\
\dot{n}_y &=& -f_y - u_x n_z\\
\dot{n}_z &=& -f_z-u_yn_x + u_xn_y \label{eq:ode_local_simp4}
\end{eqnarray}
where $x$, $y$, and $z$ are the first, second, and third component of a vector, respectively. $K_{BT,ij}$ means the entry of $K_{BT}$ at row $i$ and column $j$. Appendix \ref{Appendix A} shows the detailed procedures to obtain the ODEs (\ref{eq:ode_local_simp1}) and (\ref{eq:ode_local_simp2}). And (\ref{eq:ode_local_simp3}) - (\ref{eq:ode_local_simp4}) is the same as (\ref{eq:ode_local2}). 
Note that both the integral variables and the boundary conditions are defined in local frame, the curvature $u_x$ and $u_y$ can be solved by simple backward integration from the distal point ($s = L$) to the proximal point ($s = 0$), which requires no iterative computation to solve a standard BVP. Therefore, the computational complexity is only determined by the number of nodes we divide along the arc length. Moreover, solving (\ref{eq:ode_local_simp1}) - (\ref{eq:ode_local_simp4})  will lead to the same result as solving the model in (\ref{eq:ode1}) - (\ref{eq:ode4}), but the former method will have faster performance for curvature calculation. 

\section{Force Estimation Approach}
\subsection{Conditions For Accurate Force Estimation}
\begin{figure}[h]
    \includegraphics[width=0.99\linewidth]{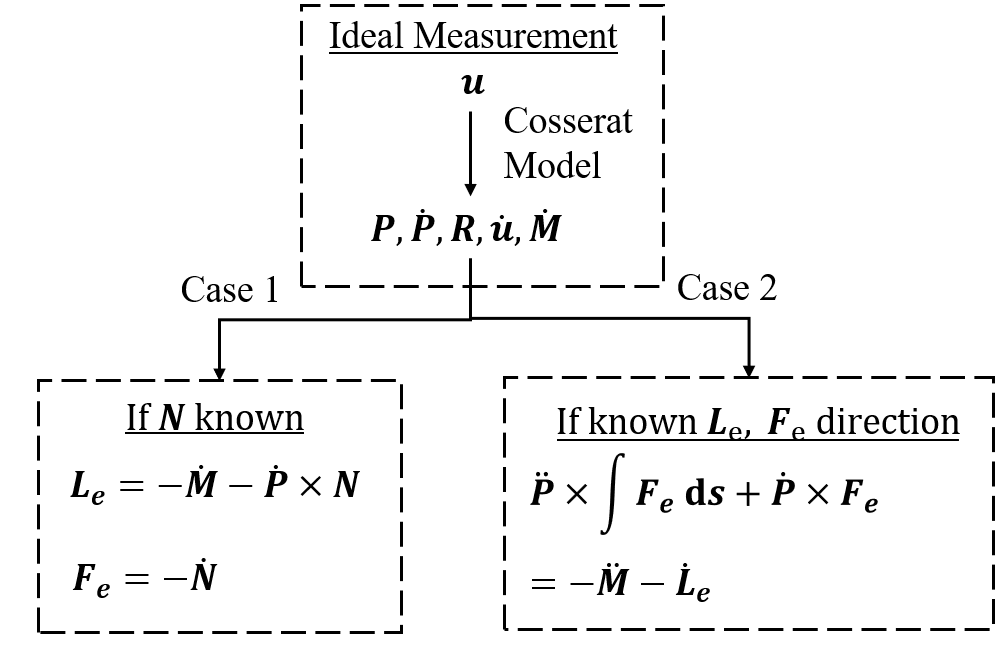}
    \caption{Two cases where $F_e$ can be estimated accurately with ideal measurements. In both cases additional information has to be provided to achieve accurate result.}
    \label{fig:Error Analysis diagram}
\end{figure}

While computing the configuration of a rod under external forces and moments can be straightforward, its inverse mechanics is ill-conditioned \cite{kabanikhin2008definitions}. Force estimation based on variables in configuration space can become inaccurate even with perfect data measurement. In this subsection, we analyze the conditions where force can be estimated accurately.
We assume the mapping between the measured curvature and estimated force is 
\begin{equation}
    \mathscr{F}(\Tilde{\B u}) =\Tilde{\B f}_e
\end{equation}
where $\mathscr{F}$ is a Cosserat model-based method to estimate the force $\Tilde{\B f}_e$ from measured curvature $\Tilde{\B u}$. The error of the estimation can be defined as 
\begin{equation}
    \mathscr{E}_\mathscr{F}(\Tilde{\B u}) = \lVert \Tilde{\B f_e} - \B f_e^{gt}\rVert
\end{equation}
where $\mathscr{E}_\mathscr{F}(\Tilde{\B u})$ is the 2-norm error of method $\mathscr{F}$ with measured $\Tilde{\B u}$, $\B f_e^{gt}$ is the ground true force. If the estimation method $\mathscr{F}$ is sufficiently accurate, we can write
\begin{equation}
\mathscr{E}_\mathscr{F}(\B u) \leq    \mathscr{E}_\mathscr{F}(\Tilde{\B u})
\end{equation}
where the $\B u$ is the ideal curvature measurement with no noise. Ideal measurements on curvature or points of shape cannot guarantee that $\mathscr{E}_\mathscr{F}(\B u) = 0$. Additional information has to be assumed, or measured in order to achieve accurate results. To derive what else information is necessary for force estimation, we assume the ideal measurements of $\B u$ is known. As shown in Fig. \ref{fig:Error Analysis diagram}, variables $\B P$, $\B R$, $\B M$ and their derivatives can be directly computed from ideal measurements using Cosserat rod model. However, because of the cross product in (\ref{eq:ode4}), unique solution of $\B F_e$ can not be achieved with only curvature measurements. Thus, we list two special cases in Fig. \ref{fig:Error Analysis diagram} that can theoretically complete the information for force estimation. 

Case 1 indicates the internal force $\B N$ need to be measured in additional to curvatures for force estimation. The application of case 1 requires multiple force sensors mounted along the length of the manipulator. Case 2 requires the knowledge of $\B L_e$ and the direction of $\B F_e$. But assumptions can be made on these two variables: 
\begin{enumerate}
    \item  No external moments ($\B L_e = \B 0$).
    \item External forces are always perpendicular to the surface of the manipulator $F_{e,z}=0$. 
\end{enumerate}
One advantage of these assumptions is that force can be estimated without any additional mounted sensors. However, these assumptions also limit the application range of force estimation. The surface of the manipulator has to be smooth, and the manipulator has to work in an environment where friction is trivial to the shape. Compared with case 1, case 2 with the aforementioned two assumptions allows force estimation without additional sensors. The goal of these paper is to estimate force only from curvature data. Thus, we will use these assumptions throughout the following paper.

\subsection{Point Forces For Cosserat Model}
\begin{figure}[h]
\centering
    \includegraphics[width=0.8\linewidth]{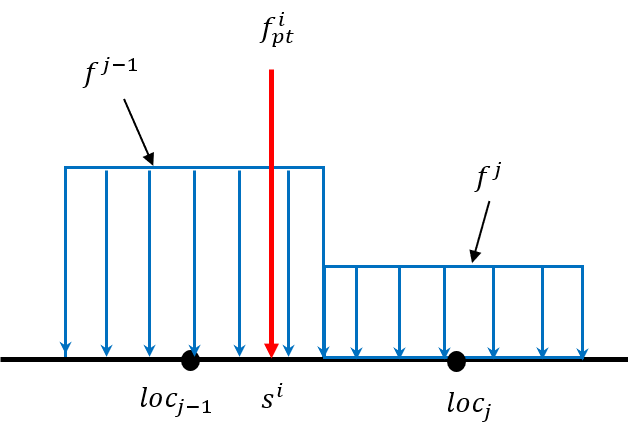}
    \caption{Distribution of point force to adjacent nodes.}
    \label{fig:point2distr}
\end{figure}

The force defined in Cosserat rod theory is distributed force whose unit is N/m. However, the objective of this paper is to estimate the magnitude and location of point force whose unit is N. Conversion has to be performed between point force and distributed force. Firstly, we define the applied point forces in a new force vector which contains the locations and magnitude components of all forces.
\begin{equation}
    \B f_{vec} = [s^1,f^1_{pt,x},f^1_{pt,y},s^2,f^2_{pt,x},f^2_{pt,y},...,s^h,f^h_{pt,x},f^h_{pt,y}]^T \label{eq:force vector}
\end{equation}
where $s^i$ is the location of the $i^{th}$ point force, $f_{pt,x}^i$ and $f_{pt,y}^i$ are the components of the $i^{th}$ point force, and $h$ is the total number of forces acting on the continuum robot. The arc length of the manipulator can be divided into $q-1$ segments, and this gives
\begin{equation}
    \B {Loc} = [loc_1, loc_2, ..., loc_q]
\end{equation}
where $loc_i$ is the location of the $i^{th}$ node, $\B {Loc}$ is the whole list of the nodes. The conversion from point forces to distributed forces is the process to distribute $\B f_{vec}$ on the nodes $\B {Loc}$. Fig. \ref{fig:point2distr} shows an example to distribute the point force $f^i_{pt}$ to two adjacent nodes $j-1$ and $j$, the point force is distributed linearly according to the arc length distance between the two nodes. Therefore, the distributed forces can be calculated by
\begin{eqnarray}
f^{j-1} &=& \frac{f^i_{pt}}{(loc_j - loc_{j-1})^2}(s^i - loc_{j-1})\\
f^{j} &=&\frac{f^i_{pt}}{(loc_j - loc_{j-1})^2}(loc_{j} - s^i)
\end{eqnarray}
In the case where multiple point forces are close and distributed to the same node, the forces on that node will be superposed. 
After knowing the distributed forces along the rod. The mechanics model is complete and curvature can then be calculated through the integration of (\ref{eq:ode_local_simp1}) - (\ref{eq:ode_local_simp4}).

\subsection{Force Estimation From Observed Curvature}
A robust method for force estimation is to minimize the least square loss between the calculated curvature with measured curvatures. 
\begin{equation}
    loss_u(\B f_{vec}) = \lVert \B u_{cal} - \Tilde{\B u}\rVert^2 \label{eq:loss}
\end{equation}
where $\B u_{cal}$ is the curvature calculated from $\B f_{vec}$, $\Tilde{\B u}$ is the curvature measured by sensors.

\begin{figure}[h]
    \includegraphics[width=0.99\linewidth]{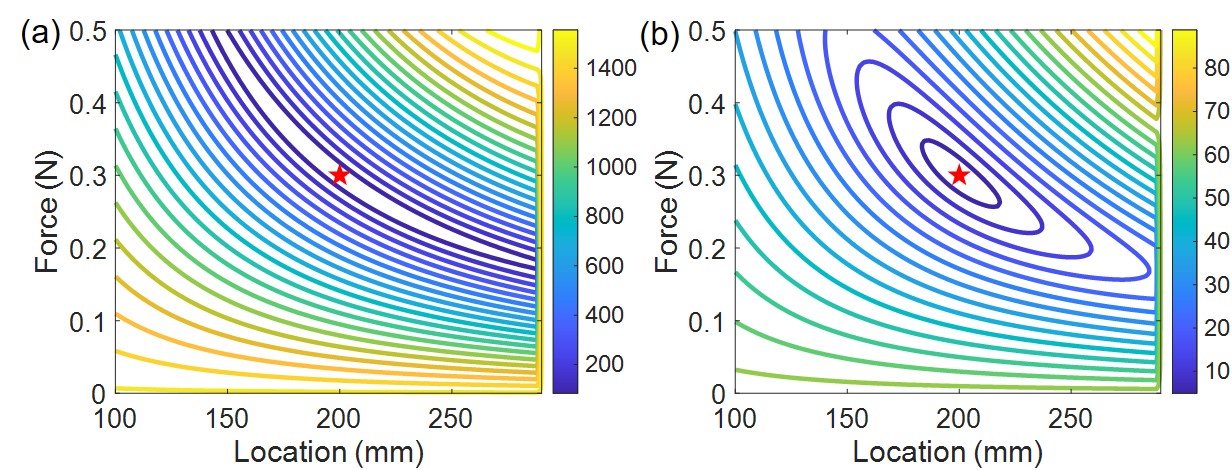}
    \caption{Loss maps computed by (a) shape (b) curvature. }
    \label{fig:Results_lossMap}
\end{figure}

Similar to minimizing the loss of curvature, the loss of shape $loss_{P}(\B f_{vec})$ could also be used for the objective function of the optimization algorithm. Fig. \ref{fig:Results_lossMap} shows an simulation example of the loss of shape $loss_{P}(\B f_{vec})$ and loss of curvature $loss_{u}(\B f_{vec})$ with various force magnitudes and locations. The loss map is calculated by following steps: 1) Choosing $\B f_{vec} = [200\quad0.3\quad0]^T$ as the ground truth, and calculate the curvature $\Tilde{\B u}$ and shape $\Tilde{\B p}$ using Cosserat rod theory. The $\Tilde{\B u}$ and $\Tilde{\B p}$ can be assumed as the measured data from sensors.  2) Calculate $\B u_{cal}$ and $\B P_{cal}$ with the same method, but use different $\B f_{vec}$ whose first component ranges from 100 - 290 mm and the second component ranges from 0 - 0.5 N. 3) use (\ref{eq:loss}) to compute the loss for each pair of force magnitude and location. As illustrated in Fig. \ref{fig:Results_lossMap}, the loss of curvature (b) shows better convex property than that of shape coordinates (a), and thus more robust results can be calculated \cite{bubeck2014convex}. 
Therefore, we use curvature based method to estimate the force. 

\section{Experiments and Results}
\subsection{Experiment Setup}
\begin{figure}[h]
    \includegraphics[width=0.99\linewidth]{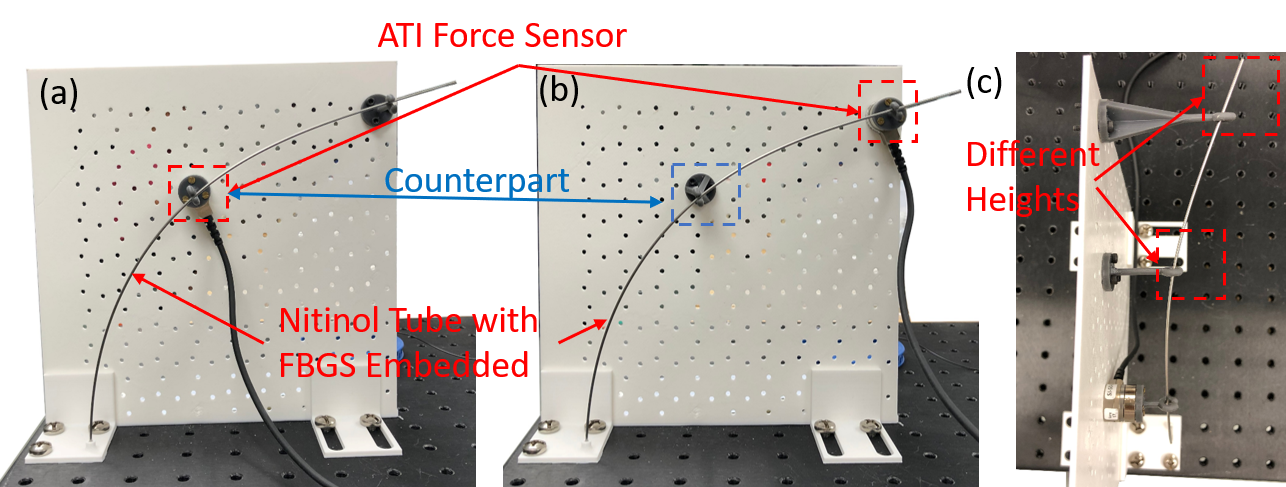}
   \caption{Experiment setup for force estimation. The ATI force sensor with a probe is mounted on a fixed vertical wall. The FBGS is inserted inside Nitinol tube for curvature measurement. Counterpart of each force sensor with probe are designed to keep the shape of Nitinol in (a) and (b) the same, and the force sensor can swap the location. Different height design of the probe in (c) allows for 3D shape deformation.}    \label{fig:Experiment setup}
\end{figure}

As illustrated in Fig. \ref{fig:Experiment setup}, the experiments were performed by adding multiple forces to a straight Nitinol tube. ATI force sensor (ATI Industrial Automation, United States) with 3D printed probe was mounted on a fixed vertical board. The probes are designed with different height in order to contact with the Nitinol tube (Fig. \ref{fig:Experiment setup}c). For every single probe that mounted on the force sensor, a counterpart probe (Fig. \ref{fig:Experiment setup}a and Fig. \ref{fig:Experiment setup}b) is also designed to keep the same distance from the probe head to the wall. This allows the force sensor with probe to be interchangeable with its counterpart such that the shape of Nitinol tube remains unchanged. For example, the Nitinol tube in Fig. \ref{fig:Experiment setup}a and Fig. \ref{fig:Experiment setup}b has the same shape, but the location of the force sensor swapped. Thus, we can use one force sensor to measure multiple external forces. Fiber bragg gratings sensors (FBGS International NV, Belgium) was inserted inside the Nitinol tube to measure the curvature. 

\subsection{Model Calibration}
The calibration of the experiment was conducted in two step: 1) location calibration 2) stiffness calibration. Since the FBGS fiber is transparent, it is hard to figure out the locations of gratings inside the fiber. But the relative locations (spacing between adjacent gratings) are specified on the user manual (20 mm). The objective of location calibration is to ensure the curvatures we measured are aligned with the positions on the Nitinol. Three cases of single force at different locations were recorded, and the bias of the location can be minimized by adding an offset to the results.
\begin{equation}
    s_{est} = s_{cal} + s_{bias}
\end{equation}
where $s_{est}$ is the estimated location of the external force, $s_{cal}$ is the result of the model, and $s_{bias}$ is the bias to offset the error. After location calibration, we conducted stiffness calibration to match the calculated force magnitude with the measured force magnitude. Notice that the stiffness change has trivial impact on the location estimation, therefore the location calibration is still valid after completing the stiffness calibration.The result of the calibration is listed in TABLE \ref{tab: calibrated parameters}.

\begin{table}[t]
\caption{Calibrated Parameters}
\centering
\begin{tabular}{llll}
\hline
Name      & Variable  & Value & Unit \\ \hline
Nitnol tube length & $loc_{tip}$   & 290 & mm   \\
Inner tube diameter & $d_{in}$   & 1.118 & mm   \\
Outer tube diameter & $d_{out}$  & 1.397 & mm   \\
Young's modulus & $E_{ni}$   & 67    & GPa  \\
Location bias & $s_{bias}$ & -3.12 & mm   \\ \hline
\end{tabular}
\label{tab: calibrated parameters}
\end{table}

\subsection{Force Number Estimation}
\begin{figure}[h]
\centering
    \includegraphics[width=0.9\linewidth]{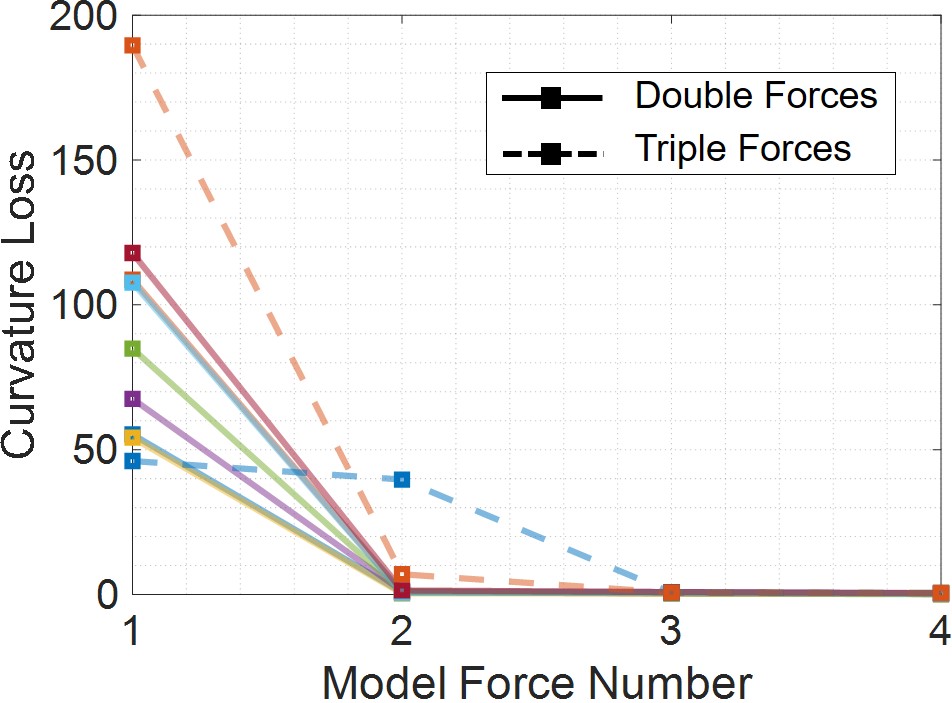}
    \caption{Curvature loss $loss_u(\B f_{vec})$ for various model force number $h$. }
   \label{fig:Results_forceNumber}
   \end{figure}

The output of the force estimation method is a force vector $\B f_{vec}$ defined in (\ref{eq:force vector}), which specifies the number of force $h$. The ideal estimation result should assign  the force number $h$ in $\B f_{vec}$ exactly the same as the real force number before running the force estimation method. In practice, knowing the number of forces on a manipulator in advance is hard through other measurements. Mismatching the model force number with real force number will reduce the estimation accuracy, even though the magnitude of the redundant forces are small. However, as shown in Fig. \ref{fig:Results_forceNumber}, the mismatch of force number can be detected by setting a threshold for curvature loss $loss_u(\B f_{vec})$. 

For the situation where $h$ is less than the real force number, it is obvious that a large curvature loss occurs. This indicates that  $h$ needs to increased in the model for better estimation. For the situation where the $h$ is larger than the real force number, the curvature loss value drops significantly. This indicates that a smaller $h$ will be assigned in the model for better estimation. In this paper, we start with $h=1$ and gradually increase the $h$ until the curvature loss drops below a specified threshold (3.0, in our case) at the first time. This is also valid for single force estimation, the curvature loss of which is below the threshold at the beginning ($h=1$). 

\subsection{Force Estimation Results}
\begin{figure}[h]
\centering
    \includegraphics[width=0.99\linewidth]{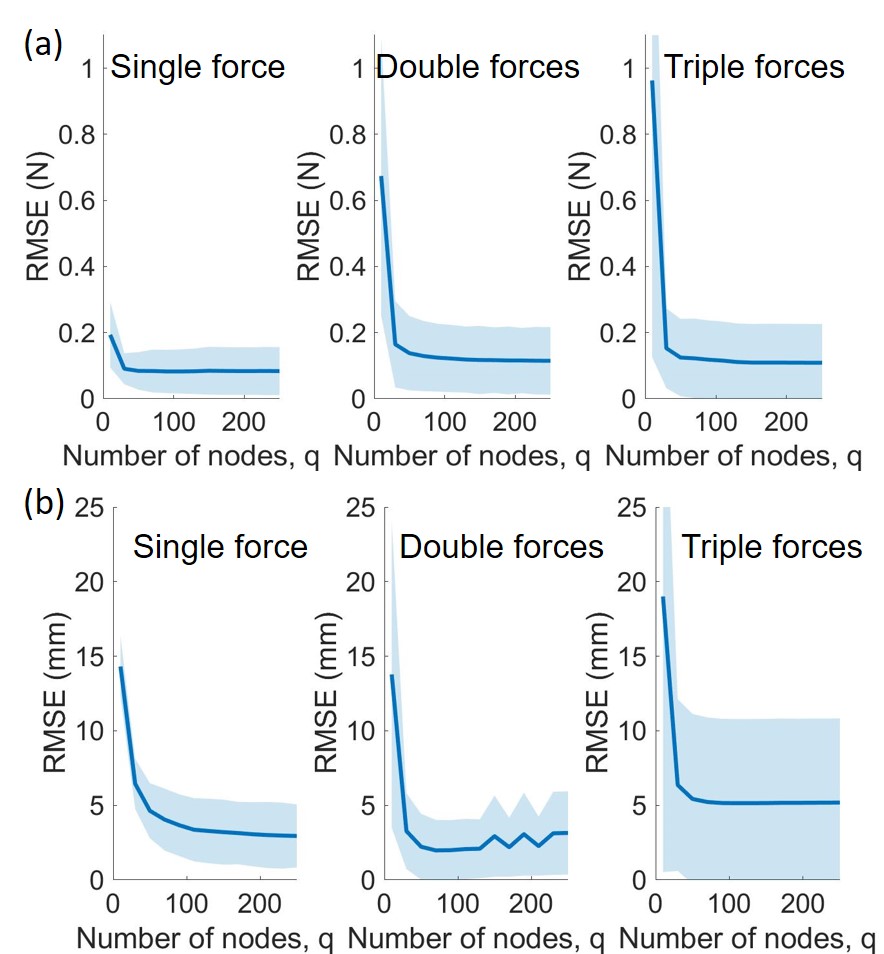}
    \caption{(a) Accuracy of the location estimation with respect to the number of nodes $q$. (b) Accuracy of the force magnitude estimation with respect to the number of nodes $q$. }
        \label{fig:accuracy with q}
\end{figure}

The error analysis of the force estimation result is grouped into 3 categories: single force estimation analysis, double forces estimation analysis, and triple forces estimation analysis. For each category, the experiment was performed 13 times by applying varying contact force (0.27 - 1.96 N) at 13 different locations on the Nitinol tube. Intuitively, the number of nodes $q$ can also have impact on the estimation results. As demonstrated in Fig. \ref{fig:accuracy with q}, the RMSE of both force magnitude and location decrease as the number of nodes $q$ increases. However, if the $q$ is sufficiently large, RMSE can barely decrease any more. 

For the force magnitude estimation with $q = 250$ in Fig. \ref{fig:accuracy with q}a, the RMSE of single force estimation, double force estimation, and triple force estimation are $0.084 \pm 0.073$ N, $0.115 \pm 0.102$ N, and $0.1090 \pm 0.1173$ N, respectively. The range of the measured force is 0.3 - 1.5 N. Therefore, the proposed method can estimate the force magnitude accurately, though larger number of forces can slightly reduce the estimation accuracy on force magnitude.
For the force location estimation with $q=250$ as shown in Fig. \ref{fig:accuracy with q}b, the RMSE of the single-location estimation, double-location estimation, and triple-location estimation is $2.95\pm2.11$ mm, $3.15\pm2.79$ mm, and $5.1873\pm5.64$ mm, respectively. This indicates that the larger location error will occur when the number of external forces increases. The worst prediction in triple force location error is 10.59 mm, which is 3.65\% of the total length ($loc_q = 290$ mm).

\begin{figure}[h]
    \includegraphics[width=0.99\linewidth]{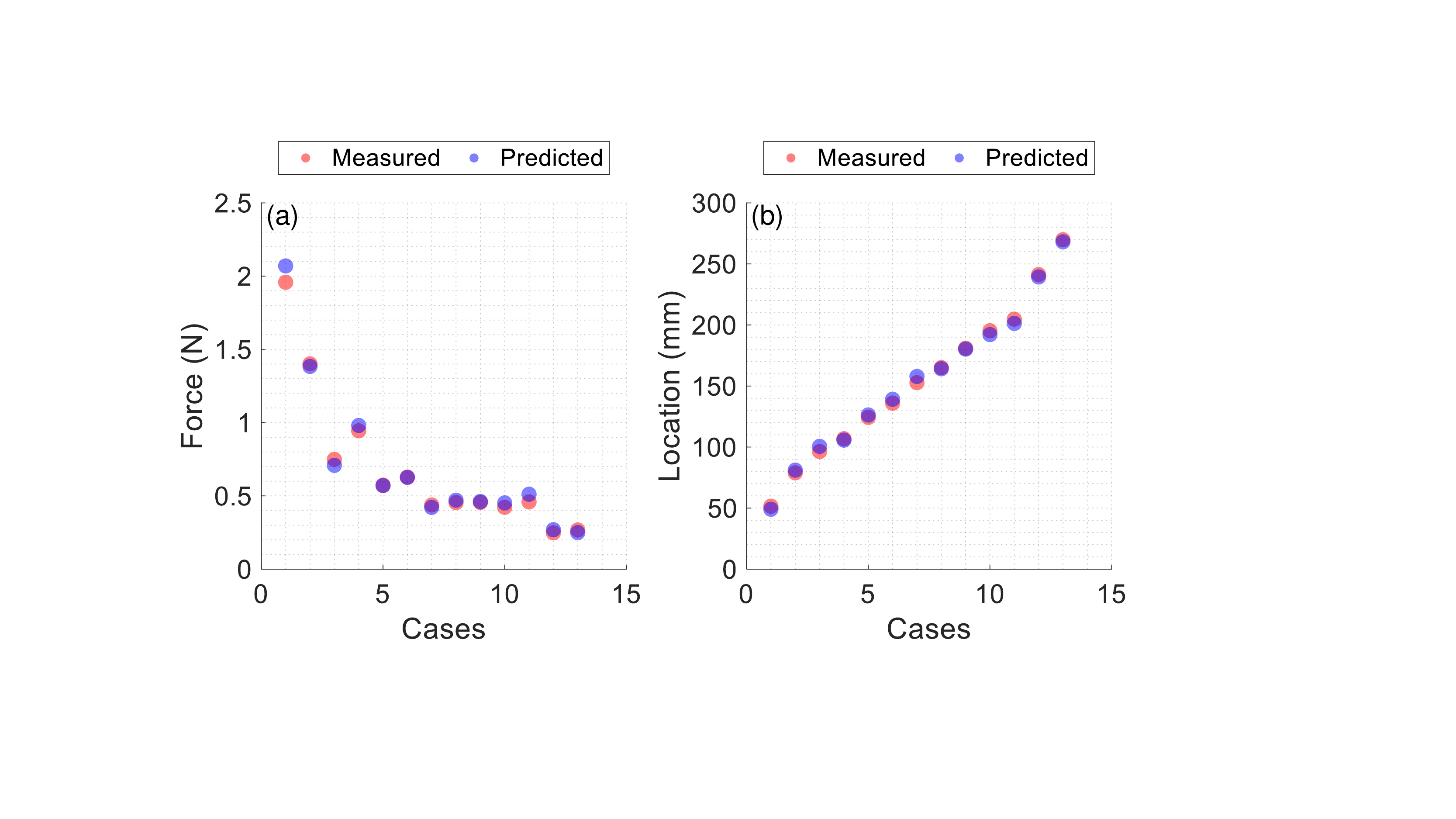}
    \caption{Experimental results for single force estimation. (a) force magnitude estimation. (b) force location estimation.}
   \label{fig:Results_singleForceEstimation}
   \end{figure}

\begin{figure}[h]
    \includegraphics[width=0.99\linewidth]{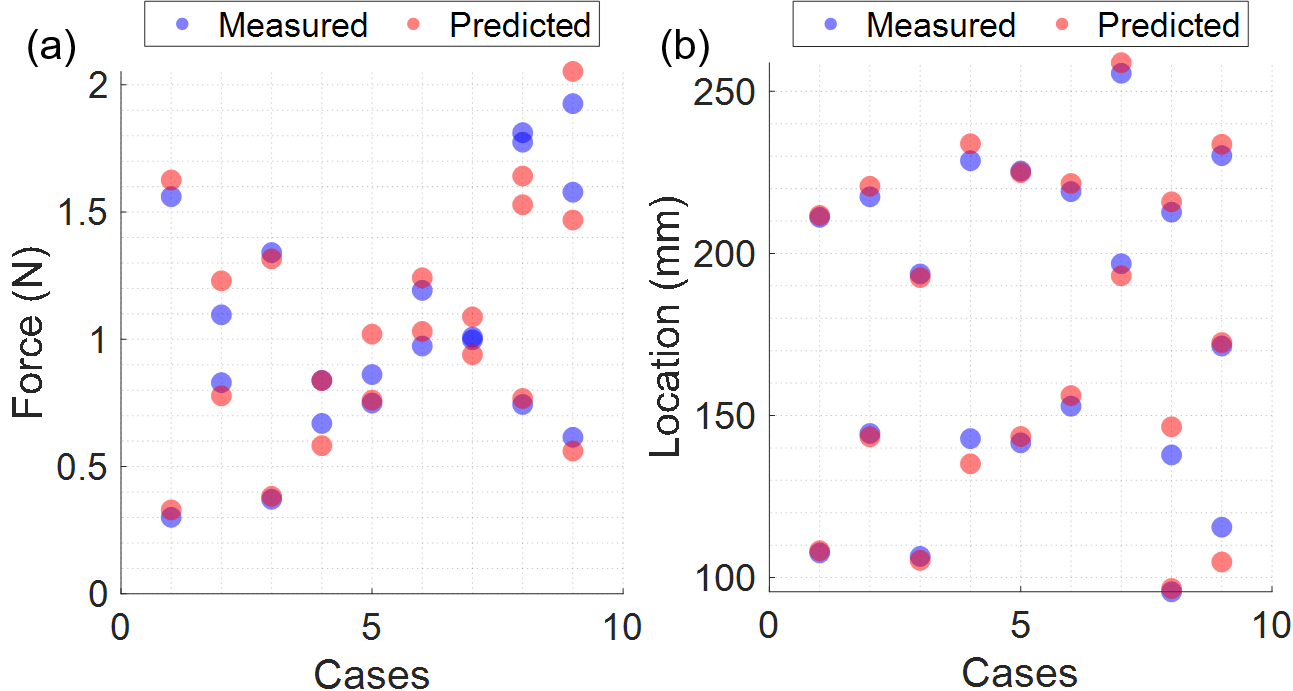}
   \caption{Experimental results for double and triple forces estimation. (a) force magnitudes estimation. (b) force locations estimation.}    \label{fig:Results_doubleForceEstimation}
\end{figure}

Fig. \ref{fig:Results_singleForceEstimation} and Fig. \ref{fig:Results_doubleForceEstimation} summarize the the force estimation results ($q = 250$) for single force and multiple forces, respectively. The blue dots are the measured result and the red dots are the predicted results. Most of the cases in Fig. \ref{fig:Results_singleForceEstimation} can be predicted accurately except the case 1 (magnitude error of 8.72\%). This error occurs because the curvature can only be measured by the first few gratings (the spacing of adjacent gratings is 20 mm), while the rest gratings will read 0 because no internal moment exists after the location where the force applied. In Fig. \ref{fig:Results_doubleForceEstimation}, each case has two or three values, which refer to the two or three forces, respectively. The red dots and the blue dots are aligned well with each other, which demonstrate the accuracy of the proposed estimation method.

\subsection{Computational Speed}
 In this section, we aim to evaluate the feasibility for realtime force estimation by analyzing the force estimation speed. 
  The laptop for the speed test has a CORE-i7 Intel CPU with 6 cores. The force estimation method was implemented in MATLAB, and the optimization problem is solved using \emph{fmincon} function with interior-point algorithm. The number of nodes $q$ as well as the number of inputs (forces and locations) also have significant impact on the calculation speed.
 We tested the method with $q$ in the range of 50 and 250. For $q$ larger than 250, the accuracy of this method will not improve, as shown in Fig. \ref{fig:accuracy with q}. Since our main motivation to increase $q$ is to ensure the accuracy of the estimation, $q$ larger than 250 will be excluded in the test. 
 
\begin{figure}[h]
    \includegraphics[width=0.99\linewidth]{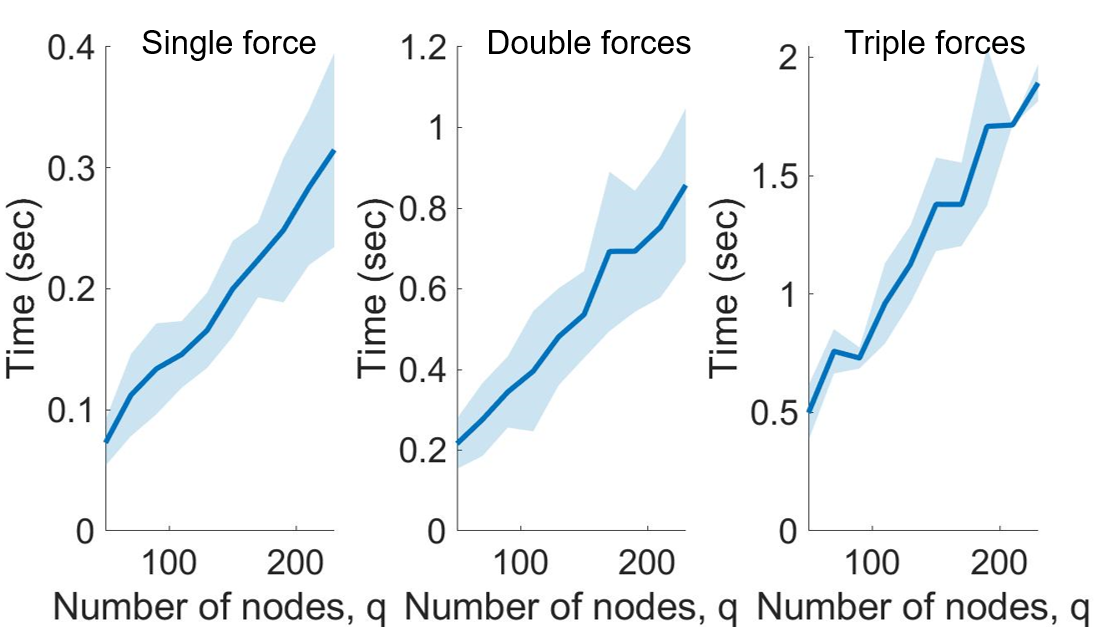}
   \caption{Force estimation time varies with the number of node $q$.}    \label{fig:Results_timeWithQ}
\end{figure}

 The consumed time for single, double, and triple forces estimation is shown in Fig. \ref{fig:Results_timeWithQ}. For best accuracy ($q = 250$), single force estimation can be completed in $0.315\pm 0.080$  s, double force estimation can be completed in $0.857\pm 0.191$ s, and the triple force estimation can be completed in $1.890 \pm 0.078$ s. For better performance, smaller $q$ which does not significantly compromise the accuracy is preferred to achieve faster force estimation. For the case $q=90$, single force estimation takes $0.134 \pm 0.038$ s, double force estimation takes $0.344\pm 0.088$ s, and triple force estimation takes $0.730 \pm 0.045$ s.

Notice that we can calculate the same curvature by solving BVP proposed in \cite{jones2009three} using Levenberg-Marquardt (LM) method, or employ the derivative propagation (DP) method proposed in \cite{rucker2011computing} to speed up the iterations. We compare the computation time for the three approaches for various $q$. TABLE \ref{tab: computation time comparison} lists out the computation time for single force estimation. Our method shows much faster computation speed than the other two methods. Typically, when $q=100$, our method is 29.0 times faster than derivative propagation method, and 101.6 times faster than Levenberg-Marquardt method. The computational speed can be further improved by implement the method in C++ language, or use better optimization algorithm if the Jacobian of the curvature loss can be computed in a faster method.

\begin{table}[t]
\caption{Computation Time Comparison}
\centering
\begin{tabular}{cccc}
\hline
Nodes      & BVP +LM  & BVP + DP  & Ours \\ \hline
$q=50$ & 4.68 s   & 2.23 s & 0.07 s   \\
$q=100$  & 14.22 s  & 4.06 s & 0.14 s   \\
$q=150$ & 21.28 s  & 4.91 s & 0.19 s   \\
$q=200$& 27.61 s & 7.63 s & 0.26 s\\
\hline
\end{tabular}
\label{tab: computation time comparison}
\end{table}
\section{Conclusions}
In this article, we analyzed the mechanics model and come up with the conditions on which the force can be estimated with only curvature measurements. After having those conditions (no external moment and no axis force such as friction), a simplified Cosserat rod theory was derived to compute the curvature in a fast speed. Least squares optimization was used to minimize the loss between computed curvatures and measured curvatures to find the optimal force vector $\B f_{vec}$, which contains the estimated force magnitudes and locations. 

The proposed method was validated on a straight 290 mm Nitinol tube, with single or multiple forces acting at different locations. The results showed that the model can estimate both the force magnitude and location accurately. The RMSE of single force estimation, double force estimation, and triple force estimation are 0.084 $\pm$ 0.073 N, 0.115 $\pm$ 0.102 N,  and  0.1090 $\pm$ 0.1173 N, respectively. The  RMSE of  the single-location  estimation,  double-location  estimation, and triple-location estimation is 2.95 $\pm$ 2.11 mm, 3.15 $\pm$ 2.79 mm,  and  5.1873 $\pm$ 5.64  mm,  respectively. Moreover, the computation time is tested, showing that the single force estimation, double force estimation, and triple force estimation can be completed in 0.134 s, 0.344 s, and 0.730 s in MATLAB R2021a. This speed is 29.0 time faster than solve BVP with DP, and 101.6 times faster than solving BVP with LM. 


\appendix
  \subsection{Deriving Curvature ODEs}
  \label{Appendix A}
In this section $K = K_{BT}$ is used for convenience. We firstly use conditions $\B u^* = \B 0$, $\B l_e = \B 0$ to simplify (\ref{eq:ode_local1}):
 \begin{equation}
     K\dot{\B u} = -\hat{\B u}K \B u + \hat{\B v}\B n \label{eq:step1}
 \end{equation}
 Expand the term $\B \hat{u}K\B u$ in (\ref{eq:step1}), 
 \begin{eqnarray}
   \hat{\B u}K\B u &=& \left[\begin{array}{ccc}
          0& -u_z & u_y  \\
          u_z & 0 & -u_x \\
          -u_y & u_x & 0 
     \end{array}\right]
     \left[\begin{array}{ccc}
         K_{11} & 0 & 0  \\
          0 & K_{22} & 0 \\
          0 & 0 & K_{33} 
     \end{array}\right]
     \left[\begin{array}{c}
         u_x\\
           u_y\\
           u_z
     \end{array}\right]\\
     &=& \left[\begin{array}{ccc}
          0& -u_z & u_y  \\
          u_z & 0 & -u_x \\
          -u_y & u_x & 0 
     \end{array}\right]
     \left[\begin{array}{c}
         u_x K_{11}\\
           u_y K_{22}\\
           u_z K_{33}
     \end{array}\right]\\
     &=&\left[\begin{array}{c}
         u_y u_z (K_{33} - K_{22})\\
           u_x u_z (K_{11} - K_{33})\\
           u_x u_y (K_{22}-K{11})
     \end{array}\right] \label{eq:step2.1}
 \end{eqnarray}
 
 For the term $\hat{\B v}\B n$ in (\ref{eq:step1}), we use Kirchhoff assumption $\B v = [0\quad 0\quad 1]^T $ and multiply all the elements
 \begin{equation}
     \hat{\B v}\B n = \left[\begin{array}{c}
           -n_y\\n_x\\0
     \end{array}\right] \label{eq:step2.2}
 \end{equation}
 For the term $K\dot{\B u}$ in (\ref{eq:step1}), we have
 \begin{equation}
     K\dot{\B u} =  \left[\begin{array}{c}
           K_{11}\dot{u}_x\\K_{22}\dot{u}_y\\K_{33}\dot{u}_z
     \end{array}\right] \label{eq:step2.3}
 \end{equation}
Now, consider the circular cross sectional condition ($K_{11} = K_{22} = K_{33}/2$) and substitute in (\ref{eq:step2.1}) will have
\begin{equation}
      \hat{\B u}K\B u = \left[\begin{array}{c}
           u_y u_z/2\\-u_x u_z/2\\0
     \end{array}\right] \label{eq:step2.4}
\end{equation}
Combine the third row in (\ref{eq:step2.2}) - (\ref{eq:step2.4}) will have 
\begin{equation}
    \dot{u}_z = 0
\end{equation}
 This means the curvature at $z$-direction will not change for this special case. Recall the boundary conditions specified in (\ref{eq:boundary_u=0}), we can conclude
 \begin{equation}
     u_z = 0\label{eq:step_uz=0}
 \end{equation}
Substitute this conclusion back to (\ref{eq:step2.4}) will obtain
\begin{equation}
    \hat{\B u}K\B u = \B 0 \label{step4}
\end{equation}
Therefore, substitute (\ref{eq:step2.2}), (\ref{eq:step2.3}), and (\ref{step4}) into (\ref{eq:step1}) will simplify the ODEs 
\begin{eqnarray}
  \dot{u}_x &=& -n_y/K_{11} \\
\dot{u}_y &=& n_x/K_{22}
\end{eqnarray}
Notice that $u_z$ is always zero and has no impact on the other variables, therefore it is excluded from the ODEs.

\bibliographystyle{IEEEtran}
\bibliography{references}
\end{document}